\documentclass{bmvc2k}

\usepackage{times}
\usepackage{epsfig}
\usepackage{graphicx}
\usepackage{amsmath}
\usepackage{amssymb}
\usepackage{booktabs,tabularx}
\usepackage[labelfont=bf]{caption}
\usepackage{xparse}
\usepackage{multirow}
\usepackage{comment}
\usepackage{subfloat}
\usepackage{bm}

\title{HS3: Learning with Proper Task Complexity in Hierarchically Supervised Semantic Segmentation}

\addauthor{Shubhankar Borse}{sborse@qti.qualcomm.com}{1}
\addauthor{Hong Cai}{hongcai@qti.qualcomm.com}{1}
\addauthor{Yizhe Zhang}{yizhe.zhang.cs@gmail.com}{2}
\addauthor{Fatih Porikli}{fporikli@qti.qualcomm.com}{1}

\addinstitution{
 Qualcomm AI Research\\
 San Diego, CA, USA\\
 \textit{Qualcomm AI Research is an initiative of Qualcomm Technologies, Inc.}
}

\addinstitution{
 Nanjing University of Science and Technology\\
 Nanjing, China\\
 \textit{Work done at Qualcomm AI Research.}
}
\runninghead{Borse et al.}{Hierarchically Supervised Semantic Segmentation}

\definecolor{sea}{rgb}{.14,.56,.14}
\definecolor{niceorange}{rgb}{.78,.44,.19}

\newcommand{\fp}[1]{\textcolor{magenta}{[\textbf{FP}: #1]}}

\begin{document}

\maketitle

\vspace{-20pt}
\begin{abstract}
\vspace{-5pt}

While deeply supervised networks are common in recent literature, they typically impose the same learning objective on all transitional layers despite their varying representation powers. 

In this paper, we propose Hierarchically Supervised Semantic Segmentation (HS3), a training scheme that supervises intermediate layers in a segmentation network to learn meaningful representations by varying task complexity. To enforce a consistent performance vs. complexity trade-off throughout the network, we derive various sets of class clusters to supervise each transitional layer of the network. Furthermore, we devise a fusion framework, HS3-Fuse, to aggregate the hierarchical features generated by these layers. This provides rich semantic contexts and further enhance the final segmentation. Extensive experiments show that our proposed HS3 scheme considerably outperforms deep supervision with no added inference cost. Our proposed HS3-Fuse framework further improves segmentation predictions and achieves state-of-the-art results on two large segmentation benchmarks: NYUD-v2 and Cityscapes. 
\end{abstract}

\vspace{-10pt}
\section{Introduction}\label{sec:intro}
\vspace{-5pt}
Aiming at labeling each pixel to a target category, semantic segmentation is a fundamental task in computer vision for various real-world applications, such as autonomous driving, AR/VR, photography, medical imaging, scene understanding, and real-time surveillance. 

Notable advancements in semantic segmentation originated with the end-to-end fully convolutional networks~\cite{long2015fully}. Researchers have since then looked extensively into various ways to further improve performance by adding different kinds of context to such networks. Some examples are the HRNet~\cite{sun2019high} branches and hierarchical multi-scale attention~\cite{tao2020hierarchical}, which add context based on scale; another similar direction is Object Contextual Representations (OCR)~\cite{yuan2019object}, which adds context related to label representations.

In these approaches, deep architectures play a key role, but at the same time, bring challenges to training. For instance, the gradients can vanish as they back-propagate through a deep network. In addition, the intermediate layers are highly unconstrained and lack prediction capability. In order to address these issues, deep supervision~\cite{lee2015deeply, wang2020deep} is recently adopted in training, where auxiliary supervisions are imposed on a few selected intermediate layers. To enable intermediate supervision, a separate segmentation head is constructed based on the features of each selected intermediate layer and supervised directly with the original ground truth annotations. In this way, the intermediate layers are tightly regularized by the target task, and more expressive gradients are generated to train the network.

Nevertheless, deep supervision neglects the fact that the intermediate layers have weaker representation powers as compared to the final layer since their features are computed by smaller sub-networks. As such, it can be highly complex for sub-networks to learn to solve the same segmentation problem as the overall network. This prevents the sub-networks from learning meaningful features, and in some cases, can even degrade the overall accuracy (as shown in Section~\ref{sec:results}). Moreover, the different representation powers of intermediate layers are not taken into account in deep supervision.

In this paper, we propose \textbf{Hierarchically Supervised Semantic Segmentation (HS3)}, the goal of which is to find the right learning task for each intermediate layer to be supervised. We attain these segmentation tasks by clustering semantic labels to form a set containing fewer classes, thus less complexity. Specifically, an earlier layer is supervised with a smaller set of classes to match the corresponding sub-network's (the part of the network up to the current layer) learning capacity. We propose a principled approach to determine the number of class clusters using a two-step training process. This approach utilizes the confusion matrices obtained after training a deep supervision baseline to perform automatic hierarchical grouping of classes. Hierarchical supervision is then applied in the second (final) training phase. We show the effectiveness of our method over deep supervision as well as over clustering based on the manual assignment using single-step training.    

\begin{figure}[t]
\centering
 \includegraphics[ width=0.7\textwidth,   keepaspectratio]{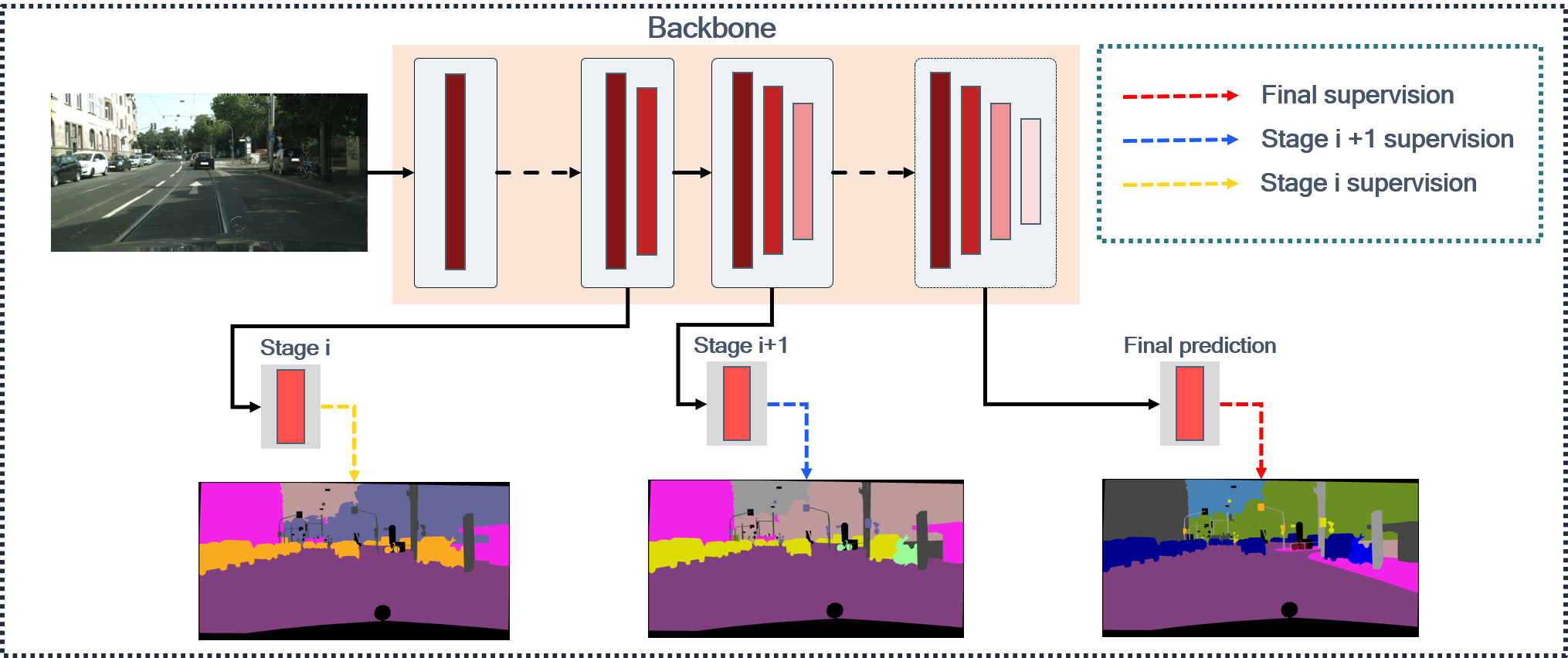}
 \vspace{-3pt}
\caption{\small \textbf{Hierarchical Supervision:} Training using our proposed HS3 scheme, where each intermediate supervision uses the right set of classes for its segmentation task, e.g., earlier layers are trained with smaller sets of classes. We show two sample intermediate stages in the figure, with $i \in \{1,\,...,\,N\}$, and the final output stage.} 
\label{fig:hs}
\vspace{-10pt}
\end{figure}

Now that each intermediate supervised layer can be trained with the suitable grouping of classes, we further propose a framework, HS3-Fuse, to fully utilize the hierarchical features generated by these layers. More specifically, we use lightweight Object Contextual Representation (OCR) modules to process the segmentation features of the supervised intermediate layers. These processed features are then aggregated and fed into the output layer to provide rich hierarchical semantic information and enhance the final segmentation performance.

The contributions of this work are summarized as follows:
\begin{itemize}
\vspace{-6pt}
\item We propose a novel hierarchical supervision scheme, HS3, for training semantic segmentation networks, which allows the supervised intermediate layers to learn with the right task complexities in terms of the sets of classes. This enhances the feature learning of the intermediate layers without incurring additional inference costs.
\vspace{-6pt}
\item We devise a novel framework, HS3-Fuse, to fully exploit the hierarchical features generated by the intermediate supervised layers. The fused features contain proper and useful hierarchical semantic context and are fed into the output layer to enhance the overall segmentation performance.
\vspace{-6pt}
\item We evaluate our proposed approach on the common benchmarks of Cityscapes, NYUD-v2 and CamVid. The results show that by utilizing HS3, we considerably improve upon the common deep supervision. HS3-Fuse then further improves the accuracy of the segmentation and achieves state-of-the-art performance.
\end{itemize}

\vspace{-10pt}
\section{Related Work}\label{sec:related}
\vspace{-5pt}

\noindent \textbf{Semantic Segmentation:} The introduction of fully convolutional networks (FCNs) paved the way for significant progress in semantic segmentation~\cite{long2015fully}. 
More recent works aim to maximize segmentation accuracy while maintaining a low inference cost, e.g., DeepLab~\cite{chen2017deeplab}, PSPNet~\cite{zhao2017pyramid}, and HRNet~\cite{wang2020deep}. Several works then build upon these backbone architectures to incorporate diverse contextual models. The added context could be based on boundaries~\cite{yuan2020segfix, Borse_2021_CVPR, takikawa2019gated}, multi-scale context~\cite{yuan2019object, tao2020hierarchical, yang2018denseaspp, lin2019zigzagnet} or relational context~\cite{yuan2019object, zhang2019acfnet, chen20182}.

\noindent \textbf{Deep Supervision:} Deep supervision was initially proposed to train classification networks~\cite{lee2015deeply, szegedy2015going} and later extended to other tasks, e.g., segmentation~\cite{wang2020deep}, depth estimation~\cite{godard2019digging}. These methods, however, assign the same task for all intermediate supervisions, ignoring the weaker learning abilities of sub-networks. Recently, \cite{li2018deep} proposes to use intermediate geometric concepts to deeply supervise a key-point estimation network. While the different capacities of intermediate layers are considered, this method is not applicable to segmentation.

\noindent \textbf{Coarse-to-Fine methods:} Some recent works apply different coarse-to-fine ideas to improve segmentation, e.g., increasing spatial resolution~\cite{lin2017refinenet, islam2017label}, mask refinement~\cite{luo2018coarse, jing2019coarse, kirillov2020pointrend}. Our method differs from these as we develop a strategy of class grouping. We use a method of matching task complexities based on sub-network capability, and hence the refinement occurs in an implicit manner.
\cite{hu2018progressive} proposes to use different sets of classes for supervision during training. However, its class grouping is manually and specifically designed for the face segmentation problem, and hence not applicable to other segmentation scenarios (e.g., driving, indoors). Furthermore, this grouping is static and cannot adapt to different networks. 
In contrast, our HS3 derives class grouping in an automated, data-driven manner, which can be applied to any segmentation application and adapt to the learning capacity of any network.

\vspace{-5pt}
\section{Hierarchically Supervised Semantic Segmentation}
\label{sec:proposed}
\vspace{-7pt}

In this section, we describe the Hierarchically Supervised Semantic Segmentation (HS3) training strategy. The first step involves identifying intermediate or transitional layers in deep networks. We use the approach illustrated by deeply supervised networks~\cite{lee2015deeply} to obtain transitional layers, which are demarcated by scale. For instance, the HRNet architecture~\cite{li2018deep} contains four stages with different scale groupings, which we identify as transitional layers. Note that our method would extend to other segmentation architectures since the identification of transitional layers can also be based on the depth of the layers. Once we identify transitional layers, we train our backbone network by imposing auxiliary supervision through segmentation heads attached to these intermediate layers. Consider a network trained with the HS3 method for $N$ stages. If $S$ is the set of ground truth predictions, we obtain $S_i$, which is a smaller set of grouped semantic labels for every stage $i, \forall i\in \{1,\,...,\,N\}$. The resulting loss function for HS3 training is given as follows:\vspace{-5pt}
\begin{equation}\label{equ:totloss}
   \mathcal{L}_\text{total} = \sum_{i=1}^{N} \gamma_i\mathcal{L}_{i}^{S_i} + \mathcal{L}_\text{final}^{S},\vspace{-7pt}
\end{equation}
where $\mathcal{L}_{i}^{S_i}$ is the segmentation loss for the $i$th intermediate supervision stage, $\gamma_i$ is the weight of the $i$th intermediate segmentation loss and $\mathcal{L}_\text{final}^{S}$ is the segmentation loss for the final network output. This approach is illustrated in Figure~\ref{fig:hs}.

Our approach differs from deeply supervised networks, for which the set of classes considered for each intermediate supervision is the same as the full set, i.e., $S_i = S$, $\forall i\in \{1,\,...,\,N\}$. However, this scheme imposes the same task complexity on all the intermediate sub-networks, in spite of their weaker and different learning capabilities. Instead, our approach supervises each intermediate layer with an optimal task complexity in terms of the set of semantic classes. We illustrate our approach to finding intermediate semantic sets next.


\begin{figure}[t]
\vspace{-0pt}
\centering
 \includegraphics[ width=0.5\textwidth,   keepaspectratio]{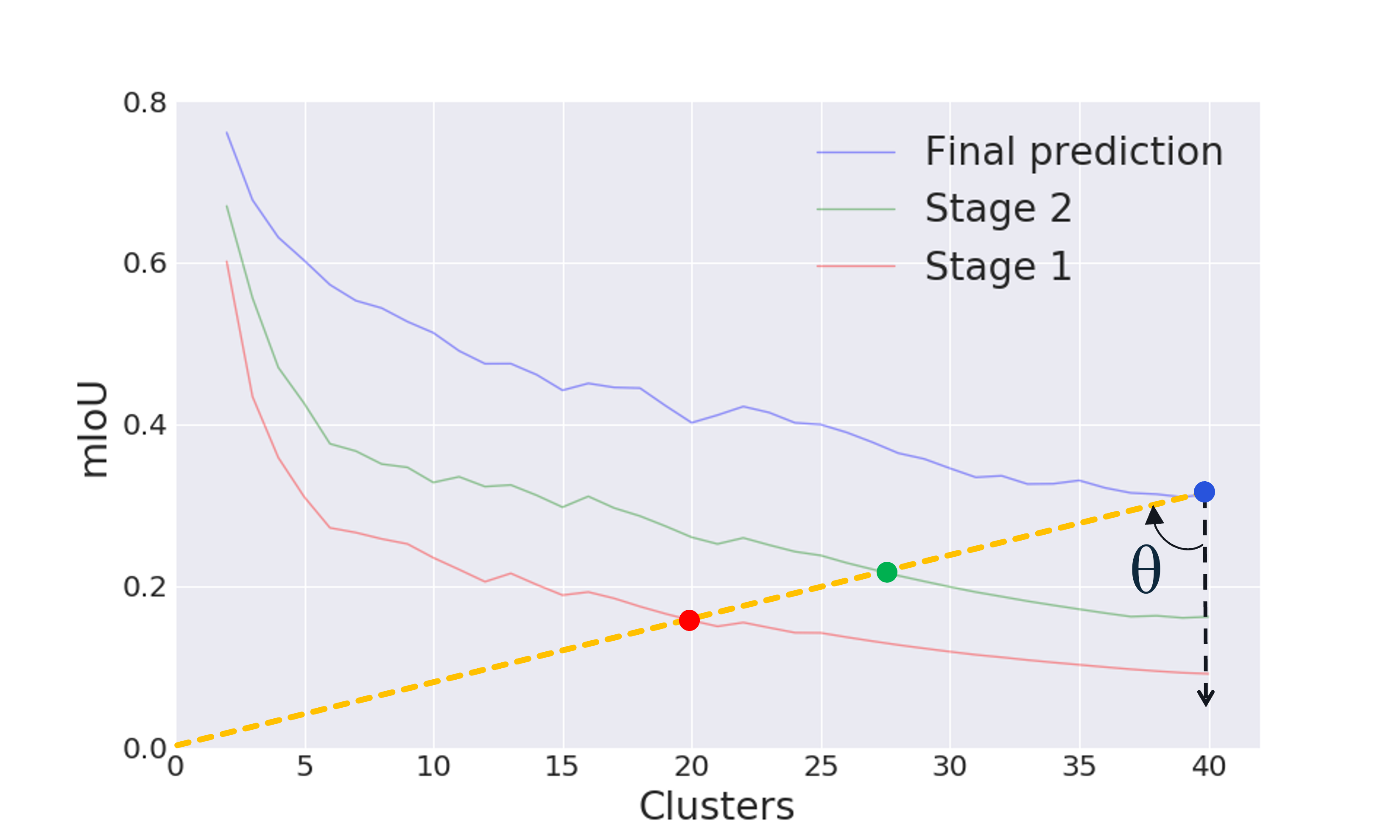}
 \vspace{-2pt}
\caption{\small \textbf{Performance-Complexity Trade-off:} We perform the analysis for an HRNetv2-w18-OCR backbone on NYUD-v2. The two intermediate layers are selected based on the scale transitions (more details in Section~\ref{sec:exp_setup}). The blue dot indicates the reference trade-off point from the final output. The red and green dots indicate trade-off points for the first and second intermediate supervision stages respectively. The x-axis shows the number of classes after clustering. 
}
\label{fig:iou_hs_rule}
\vspace{-10pt}
\end{figure}

\vspace{-5pt}
\subsection{Redefining Segmentation Tasks: Learning with the Right Classes}
\label{sec:right_classes}
\vspace{-5pt}
When applying auxiliary supervisions to a deep segmentation network, we allow each supervised intermediate layer to perform a segmentation task that is of the right complexity to it, in terms of the set of classes. We show in this part how to determine this right complexity by analyzing the trade-off between task performance and task complexity.

\vspace{-5pt}
\subsubsection{Segmentation Accuracy vs. Segmentation Complexity} 
\vspace{-5pt}

In order to understand the capabilities of the intermediate layers, we first perform a study on the segmentation performance as a function of the task complexity for each of these layers. From the available training data, we reserve a small subset as an \textit{analysis} set and use the rest as a \textit{reduced training} set.\footnote{For instance, in the case of the Cityscapes dataset, we use 90\% of the training data as the \textit{reduced training} set and the remaining 10\% as \textit{analysis} set. Note that these sets are always disjoint from the validation/test data.} First, we train the full segmentation network using vanilla (existing) deep supervision on the \textit{reduced training} set, where all the intermediate supervision stages use the full set of classes. Once the network is trained, we compute the confusion matrix, $C_i$, for each supervised intermediate layer $i$, as well as for the final layer, based on the \textit{analysis} set.

Next, we study how the segmentation accuracy varies as a function of the number of classes. This indicates the task complexity. Given a target number of classes, we apply spectral clustering~\cite{von2007tutorial} to the full set of classes based on an affinity matrix $A_i = (C_i + C_i^T)/2$, which is the symmetric version of the confusion matrix, for each intermediate supervision stage. As the clustering algorithm merges similar classes (e.g., person and rider), we are able to obtain sets of classes with sizes from $2$ to $K-1$, where $K$ is the number of classes in the full set. Note that for any two intermediate supervision stages, the sets of classes can be different even when their sizes are the same.

Based on these reduced sets of classes, we re-evaluate the segmentation accuracy for each intermediate stage in terms of mean Intersection-over-Union (mIoU). This analysis reveals the trade-off between segmentation accuracy and segmentation task complexity for each intermediate stage, as well as for the final output layer. Figure~\ref{fig:iou_hs_rule} shows the trade-off analysis for an HRNetv2-w18-OCR network on NYUD-v2. It can be seen that the accuracy reduces as the task complexity increases (in terms of the number/set of classes) and that an earlier intermediate layer shows weaker capability as compared to a later layer.

\vspace{-5pt}
\subsubsection{Choosing Proper Task Complexity}
\label{sec:right_task_complexity}
\vspace{-5pt}

If the learning task is either too complex or too simple, intermediate layers will not be able to generate useful features to aid the final segmentation. To address this, we utilize the performance-complexity trade-off considering the final output as a reference and enforce the same trade-off across all the intermediate supervision stages. More specifically, we quantify this  
trade-off using the ratio between the segmentation mIoU and the number of classes.  Then, for intermediate supervision stages, we find the trade-off points that match the ratio of the reference point, as highlighted by the green and red dots in Figure~\ref{fig:iou_hs_rule}.

Once the trade-off points are identified, we can readily determine the corresponding numbers of classes, as well as the sets of classes (based on spectral clustering) for the intermediate supervisions. These sets of classes are then used to construct the segmentation losses for the respective auxiliary supervision stages in Eq.~\ref{equ:totloss}. We then train the network on the full training set, using the total loss derived from our proposed hierarchical supervisions, which produces the final segmentation model.

In Figure~\ref{fig:iou_hs_rule}, it can be seen that our proposed approach of enforcing consistent performance-complexity trade-offs can be represented by a line through the origin and the reference point (blue dot). We denote the angle between this line and the vertical line through the reference point by $\theta$. By changing $\theta$, one can adjust the trade-offs across the layers. For instance, a larger (smaller) $\theta$ places more emphasis on task accuracy (task complexity). In particular, deep supervision corresponds to setting $\theta=0^\circ$, which requires all the intermediate layers to work on the full segmentation task. As shown in Section~\ref{sec:ablation}, our proposed approach of enforcing consistent trade-offs achieves a performance very close to the optimum.

\vspace{-5pt}
\subsubsection{Using Other Clustering Methods}
\label{sec:alternative_clustering}
\vspace{-5pt}

Our proposed HS3 framework is general and can be used with any clustering algorithm, as shown in Section~\ref{sec:ablation}. For instance, instead of running \textbf{spectral clustering} on the confusion matrix, one can perform \textbf{k-means clustering} based on the features generated by a supervised intermediate layer. This also allows us to analyze the performance-complexity trade-off for each layer, where the merging of the semantic classes is conducted via k-means.

When using spectral clustering or k-means clustering, a two-phase training process is required. It is also possible to train the network only once within our HS3 framework, by utilizing a non-data-driven approach to determine the set/number of classes for each intermediate supervision stage. For instance, we can utilize human intuition to manually cluster similar classes and derive reduced sets for the intermediate layers. Another possible way is to set a constant reduction ratio of the number of classes across the layers. For instance, we set the number of classes for each intermediate stage to be $1/2$ of that in the next stage and apply \textbf{manual clustering} based on the given number of classes.

\vspace{-5pt}
\subsection{Fusing Hierarchical Features}
\label{sec:ocrfuse}
\vspace{-5pt}

\begin{figure}[t]
\centering
 \includegraphics[ width=0.7\textwidth,   keepaspectratio]{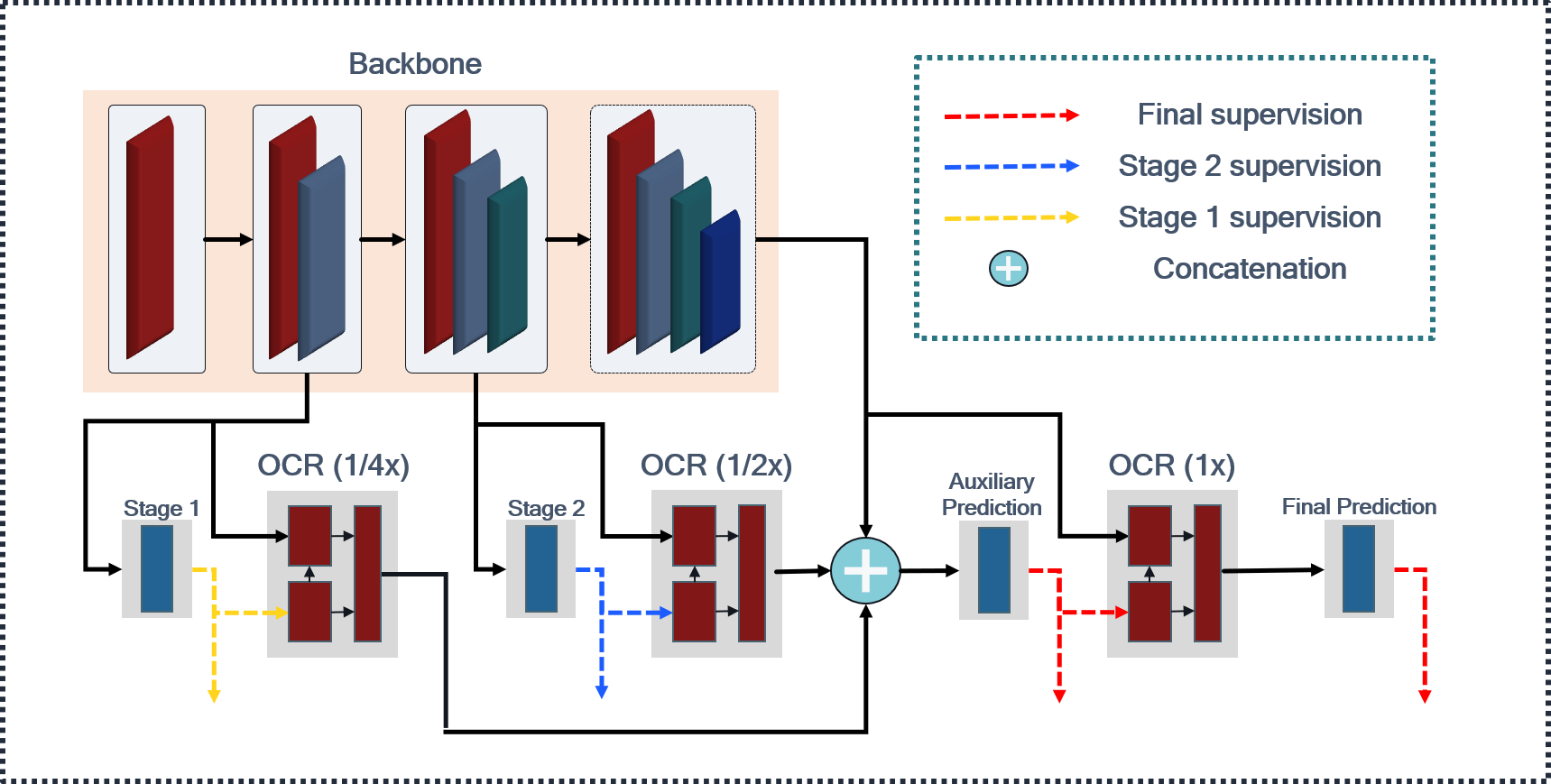}
\caption{\small \textbf{HS3-Fuse:} Using the OCR Segmentation Transformer~\cite{yuan2019object} to fuse hierarchical features back into the network.} \label{fig:hsfuse}
\vspace{-8pt}
\end{figure}

By utilizing our proposed HS3 approach, the intermediate layers can learn with the right sets of classes, which allows them to generate features of hierarchical semantic contexts at no additional computational cost. We design a fusion framework for aggregating these features to provide richer semantic information to the final segmentation. More specifically, for each intermediate supervision, we feed the segmentation features into an Object Contextual Representation (OCR) block~\cite{yuan2019object}, which enhances the features via relational context attention. These enhanced intermediate features are then fused and provided to the final segmentation layer. To reduce computational cost with the task complexity, we set the number of channels in an intermediate OCR block to be $1/2$ of that in the immediate next stage. As we shall see in Section~\ref{sec:results}, our proposed HS3 and feature fusion allow us to outperform state-of-the-art methods considerably. We illustrate the fusion process in Figure~\ref{fig:hsfuse} for the case of two intermediate supervision stages. We refer to combining HS3 and feature fusion as \textbf{HS3-Fuse}.


\vspace{-8pt}
\section{Experiments}
\label{sec:results}
\vspace{-5pt}
In this section (and in the supplementary file), we present extensive performance evaluations of our proposed approach. We compare HS3 and HS3-Fuse methods with their baseline networks, deep supervision, as well as the latest state of the art. We further conduct ablation studies on our proposed approach.

\vspace{-8pt}
\subsection{Experimental Setup}\label{sec:exp_setup}
\vspace{-5pt}
\noindent \textbf{Datasets:} We analyze semantic segmentation performance on two datasets, NYU-Depth-v2 (NYUD-v2)~\cite{Silberman:ECCV12} and Cityscapes~\cite{cordts2016cityscapes}. We use the original 795 training and 654 testing images for NYUD-v2. We further split the training set into 695 \textit{reduced training} samples and 100 \textit{analysis} samples. For Cityscapes, we use their 2975/500/1525 \textit{train}/\textit{val}/\textit{test} splits to report performance. We further split the training set into 2675 \textit{reduced training} samples and 300 \textit{analysis} samples. Models reported on \textit{test} set are trained using \textit{train+val} set. 

\noindent \textbf{Metrics:} Our primary metric for measuring performance is the mean Intersection-over-Union (mIoU). We also show the mean Pixel Accuracy for our results on NYUD-v2, and the instance IoU (iIoU) for results on Cityscapes. We use GMAC (Multiply-Accumulative Operations in $10^9$) to measure computation cost.

\noindent \textbf{Networks:} On NYUD-v2, we use HRNetv2-w48~\cite{wang2020deep}, HRNetv2-w18~\cite{wang2020deep}, HRNetv2-w18-OCR~\cite{wang2020deep, yuan2019object}, and SA-Gate-ResNet-101~\cite{chen2020bidirectional}. On Cityscapes, we use HRNetv2-w18, HRNetv2-w18-OCR, HRNetv2-w48-OCR, and DeepLab-v3+~\cite{chen2017deeplab} with WideResNet-38 (WRN-38)~\cite{zagoruyko2016wide} backbone. We apply the intermediate supervisions to layers which transition in scale. For instance, for HRNet backbones, we attach intermediate segmentation heads to the outputs of stages 2 and 3 (and stage 4 generates the final output)~\cite{wang2020deep}.

\noindent \textbf{Training:} When applying HS3 for training, we select the sets of classes for the intermediate supervision based on our trade-off analysis and spectral clustering in Section~\ref{sec:right_classes}. More details on the training and hyperparameters can be found in supplementary materials.

\vspace{-5pt}
\subsection{Results on NYUD-v2}
\vspace{-5pt}
We report results on the NYUD-v2 validation set. As shown in Table~\ref{tab:nyud_exp}, training with our proposed HS3 method consistently improves the performance as compared to deep supervision and the baseline of no intermediate supervision. For HRNet-w48, we observed that deep supervision could even degrade the segmentation performance compared to baseline. Our fusion framework is not used in this comparison. As such, our HS3 approach improves the segmentation accuracy without incurring extra computation cost at inference.


Next, we incorporate the hierarchical predictions into the proposed HS3-Fuse framework. More specifically, we use an SA-Gate-ResNet101 backbone with the proposed fusion unit discussed in Section~\ref{sec:ocrfuse}. As shown in Table~\ref{tab:nyud_sota}, our segmentation performance is $1.2\%$ mIoU more than the SA-Gates baseline. Our HS3-Fuse also achieves better performance when comparing to the latest SOTA on NYUD-v2 using RGB-D inputs, such as InverseForm~\cite{Borse_2021_CVPR}. Furthermore, we evaluate both single-scale and multi-scale inference schemes (as proposed by~\cite{chen2020bidirectional}) for mIoU and pixel accuracy. Overall, the results indicate that our proposed approach consistently improves segmentation performance in different settings and sets the new SOTA score on NYUD-v2.

\begin{table}[t]
\small
\centering
  \begin{tabular}{c |c c| c| c} 
 \hline
 Network & DS & HS3 & mIoU & GMACs (Inference) \\ 
 \hline
 \hline
  HRNetv2-w18-OCR & & & 40.6 & 22 \\
  HRNetv2-w18-OCR & \checkmark & & 41.2 & 22 \\
  HRNetv2-w18-OCR & & \checkmark & \textbf{41.7} & 22 \\
  \hline
  HRNetv2-w48 & & & 47.2 & 110 \\
  HRNetv2-w48 & \checkmark & & 47.0 & 110 \\
  HRNetv2-w48 & & \checkmark & \textbf{47.6} & 110 \\
  \hline
\end{tabular}

\caption{\small \textbf{On NYUD-v2:} Training with the proposed Hierarchical Supervision (HS3) scheme improves performance as compared to various baselines, and also outperforms the Deep Supervision (DS) approach. The improvements come with no added inference cost.}\label{tab:nyud_exp}
\vspace{-10pt}
\end{table}

\begin{table}[t]
\small
\centering
  \begin{tabular}{c |c |c| cc} 
 \hline
 Network & Backbone & Multi-Scale Inference & mIoU & Pixel-acc \\ 
 \hline
 \hline
  CEN-RefineNet~\cite{wang2020deep} & ResNet-152 & & 51.1 & - \\
  SA-Gate~\cite{chen2020bidirectional} & ResNet-101 & & 51.5 & 76.8 \\ 
  InverseForm~\cite{Borse_2021_CVPR} & ResNet-101 & & 51.9 & 77.1\\
  HS3-Fuse (ours) & ResNet-101 & & \textbf{52.2} & \textbf{77.4}\\
  \hline
  
  Malleable 2.5D~\cite{xing2020malleable} & ResNet-101 & \checkmark & 50.9 & 76.9 \\ 
  SA-Gate~\cite{chen2020bidirectional} & ResNet-101 & \checkmark & 52.4 & 77.9 \\ 
  NANet~\cite{zhang2021non} & ResNet-101 & \checkmark & 52.3 & 77.9 \\ 
  CEN-PSPNet~\cite{wang2020deep} & ResNet-152 & \checkmark & 52.5 & 77.7 \\ 
  InverseForm~\cite{Borse_2021_CVPR} & ResNet-101 & \checkmark & 53.1 & 78.1\\
  HS3-Fuse (ours) & ResNet-101 & \checkmark & \textbf{53.5} & \textbf{78.3}\\
  \hline
  
\end{tabular}

\caption{\small \textbf{On NYUD-v2}: Comparison with recent state-of-the art RGB-D methods, both with single scale and multi-scale inference. Our proposed HS3-Fuse architecture with a SA-Gates backbone outperforms all other backbones.}\label{tab:nyud_sota}
\vspace{-10pt}
\end{table}

\begin{figure}[h]
\vspace{-2pt}
\centering
 \includegraphics[ width=0.55\textwidth,   keepaspectratio]{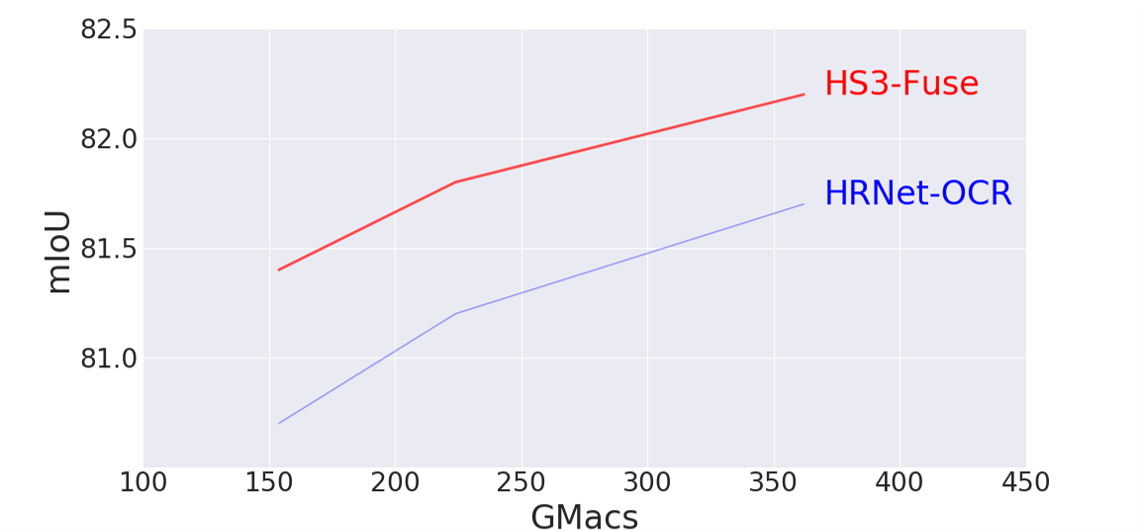}
 \vspace{-2pt}
\caption{\small \textbf{On Cityscapes \textit{val}:} Analyzing mIoU v/s GMACs performance with and without the proposed HS-Fuse architecture. We use a backbone HRNetv2-w18-OCR model and tune the OCR parameters to equalize GMAC costs.} \label{fig:cityscapes_hsfuse}
\vspace{-10pt}
\end{figure}

\vspace{-5pt}
\subsection{Results on Cityscapes}\label{sec:cityscapes_results}
\vspace{-5pt}
We provide results on Cityscapes \textit{val} and \textit{test} splits. The results on val with several backbones are summarized in Table~\ref{tab:cityscapes_exp}. We also perform inference by domain adaptation to CamVid dataset~\cite{brostow2009semantic} using the same weights, as shown in the supplementary file.
By using our proposed HS3 training scheme, we are able to consistently improve baseline scores as compared to deep supervision. These improvements come with no added inference cost. We further use our proposed HS3-Fuse approach to fully utilize the hierarchical semantic features for the case of HRNetv2-w48-OCR. We achieve an improvement in performance but with additional computational cost during inference. Hence, we also show a lighter version of HS3-Fuse by reducing the number of channels in all OCR modules by a constant factor, such that we match the GMACs required by the baseline model. It can be seen in~Table~\ref{tab:cityscapes_exp} that with the same inference cost, our lighter HS3-Fuse still considerably outperforms the baseline HRNetv2-w18-OCR. Using this technique of scaling OCR channels, we measure performance v/s computational cost at various operating points. As shown in Figure~\ref{fig:cityscapes_hsfuse}, when using the same amount of computation, our proposed approach significantly outperforms the baseline since HS3-Fuse provides richer hierarchical semantic information by fusing our extracted intermediate features.

\begin{table}[t]
\vspace{4pt}
\small
\centering
  \begin{tabular}{c c |c c| c| c} 
 \hline
 Network & Backbone & DS & HS3 & mIoU & GMACs \\ 
 \hline
 \hline
  DeepLab-v3+ & WRN38 & \checkmark & & 82.8 & 5.8K  \\
  DeepLab-v3+ & WRN38 & & \checkmark & \textbf{83.1} & 5.8K \\
  \hline
  HRNetv2-w18 & HRNetv2-w18 & & & 77.6 & 76 \\
  HRNetv2-w18 & HRNetv2-w18 & \checkmark & & 77.7 & 76 \\
  HRNetv2-w18 & HRNetv2-w18 & & \checkmark & \textbf{78.1} & 76 \\
  \hline
  HRNetv2-w18-OCR & HRNetv2-w18 & & & 80.7 & 154 \\
  HS3-Fuse & HRNetv2-w18 & & \checkmark & \textbf{81.8} & 224 \\
  HS3-Fuse (Lighter)  & HRNetv2-w18 & & \checkmark & 81.4 & \textbf{154} \\
  \hline
\end{tabular}

\caption{\small \textbf{On Cityscapes \textit{val}:} Training with the proposed Hierarchical Supervision (HS3) method improves performance compared to various baselines, and also outperforms the Deep Supervision (DS) approach with no added inference cost.}\label{tab:cityscapes_exp}
\vspace{-14pt}
\end{table}

To evaluate on the test-set, we upload predictions to Cityscapes benchmark server. We use the HS3-Fuse architecture trained using an HRNetv2-w48~\cite{li2018deep} with OCR~\cite{yuan2019object} and Hierarchical Multi-scale attention(HMS)~\cite{tao2020hierarchical} model as backbone. As seen in Table~\ref{tab:cityscapes_sota}, We achieve a gain of \textbf{0.3} mIoU and \textbf{1.7} iIoU over this baseline. We also outperform the previous state-of-the-art model(InverseForm~\cite{Borse_2021_CVPR}) by a margin of \textbf{0.1} mIoU and \textbf{0.4} iIoU. Our model ranks top in both categories among published results. We also show visual results comparing our approach to these methods in Figure~\ref{fig:city_test_qual}. Details on the predictions obtained from other methods are mentioned in the supplementary file.

\begin{table}[htb]
\vspace{4pt}
\small
\centering
  \begin{tabular}{c| c| c c} 
 \hline
 Method & Backbone & mIoU & iIoU \\ 
 \hline
  \hline
  SegFix & HRNet48-OCR & 84.5 & 65.9 \\
  Panoptic-DeepLab & Scaled WideResNet & 85.1 & 71.2 \\
  Naive Student & WideResNet41 & 85.2 & 68.8 \\
  Densely-Connected NAS & DCNAS-ASPP & 85.3 & 70.0 \\
  Hierarcical Multi-scale attention & HRNet48-OCR-HMS & 85.4 & 70.4 \\
  InverseForm & HRNet48-OCR-HMS & 85.6 & 71.4 \\
 \hline
  \textbf{HS3-Fuse(Ours)} & \textbf{HRNet48-OCR-HMS} & \textbf{85.7} & \textbf{71.7} \\
 \hline
  
\end{tabular}

\caption{\small \textbf{On Cityscapes \textit{test}:} Training with the proposed Hierarchical Supervision (HS3) framework achieves state-of-the-art scores among published methods on the live benchmark.}\label{tab:cityscapes_sota}
\vspace{-14pt}
\end{table}

\begin{figure}[htb]
\center
\begin{tabularx}{\textwidth}{c c c c}
\hspace{7mm}
Input & SegFix~\cite{yuan2020segfix} & HMS~\cite{tao2020hierarchical} & Ours\\
\hspace{8mm}
\includegraphics[ width=0.20\linewidth, keepaspectratio]{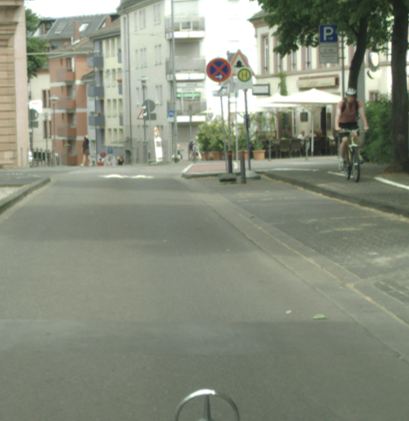} \hspace{-4mm} &
\includegraphics[ width=0.20\linewidth, keepaspectratio]{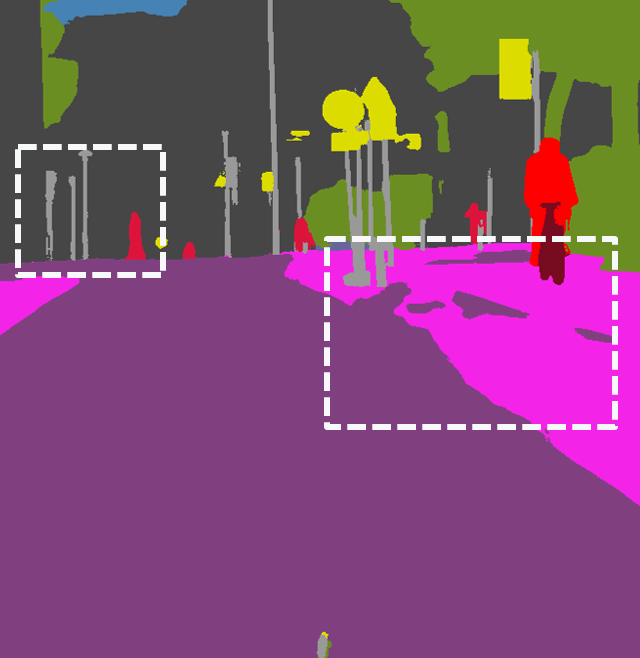} \hspace{-4mm} &
\includegraphics[ width=0.20\linewidth, keepaspectratio]{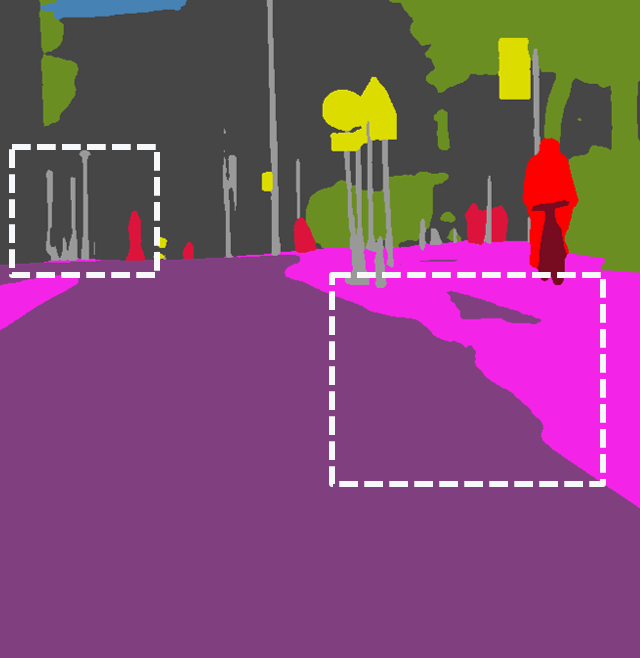} \hspace{-4mm} &
\includegraphics[ width=0.20\linewidth, keepaspectratio]{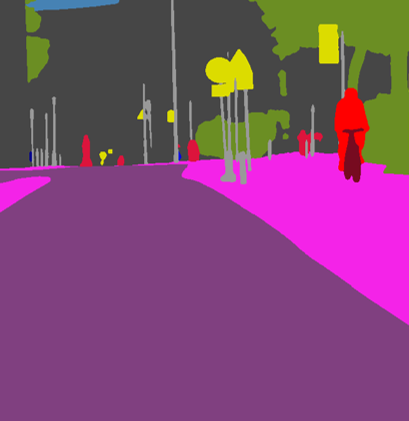} \hspace{-4mm}\\
\hspace{8mm}
\includegraphics[ width=0.20\linewidth, keepaspectratio]{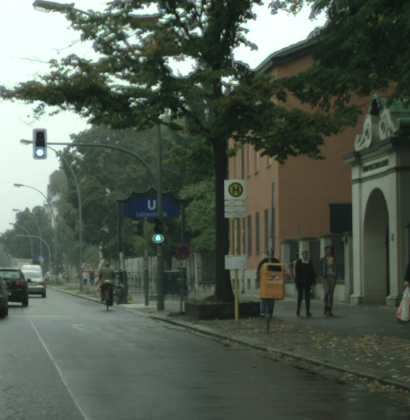} \hspace{-4mm} &
\includegraphics[ width=0.20\linewidth, keepaspectratio]{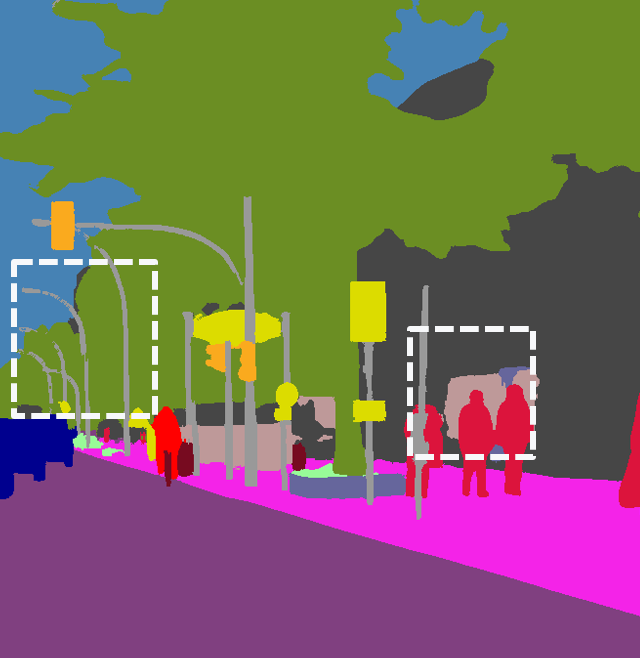} \hspace{-4mm} &
\includegraphics[ width=0.20\linewidth, keepaspectratio]{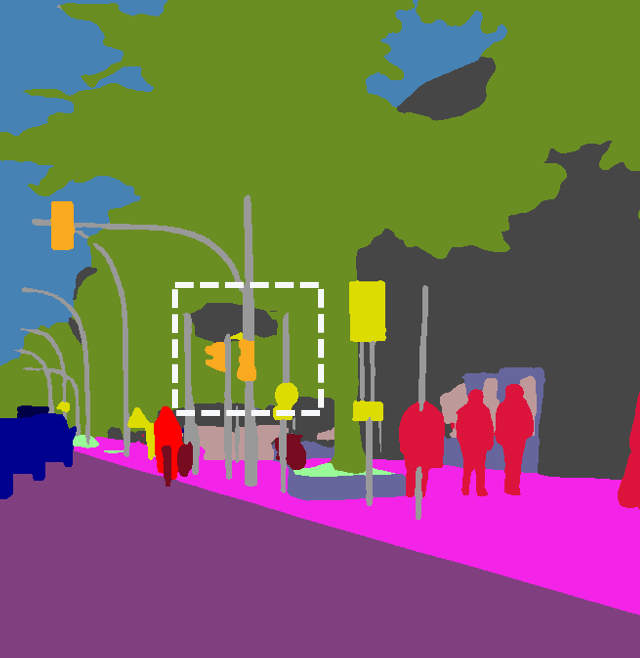} \hspace{-4mm} &
\includegraphics[ width=0.20\linewidth, keepaspectratio]{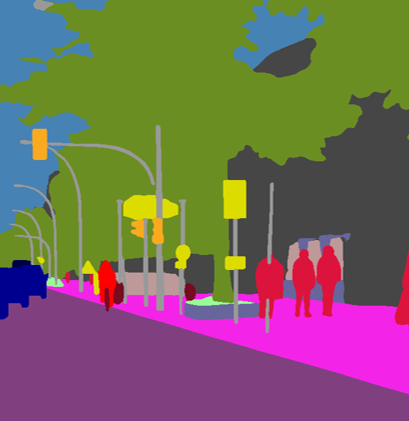} \hspace{-4mm}\\
\hspace{8mm}
\includegraphics[ width=0.20\linewidth, keepaspectratio]{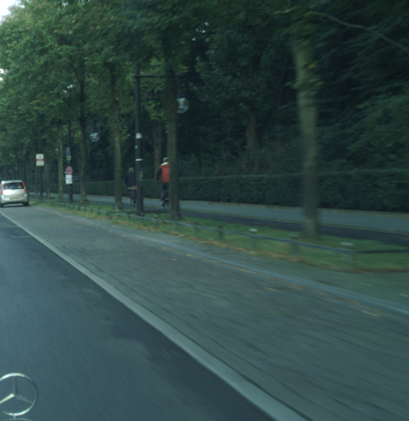} \hspace{-4mm} &
\includegraphics[ width=0.20\linewidth, keepaspectratio]{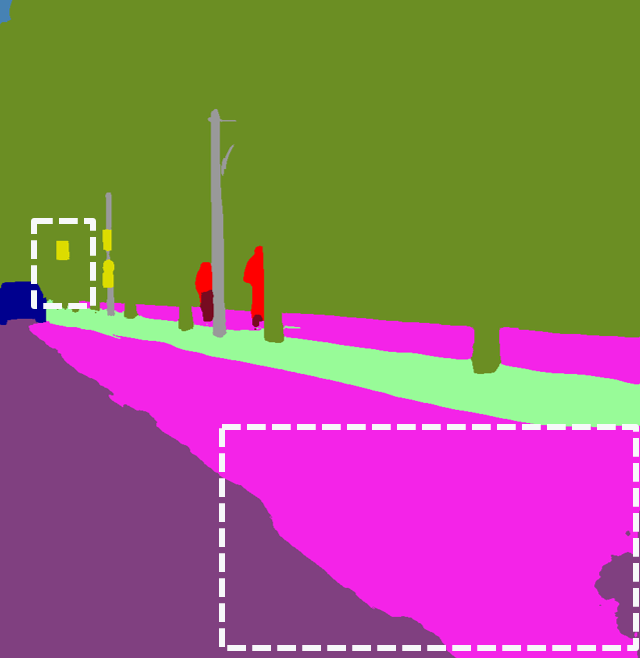} \hspace{-4mm} &
\includegraphics[ width=0.20\linewidth, keepaspectratio]{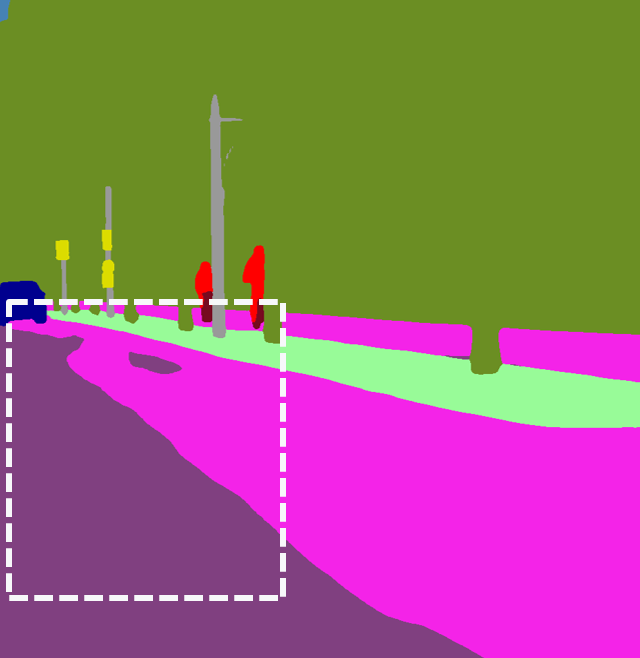} \hspace{-4mm} &
\includegraphics[ width=0.20\linewidth, keepaspectratio]{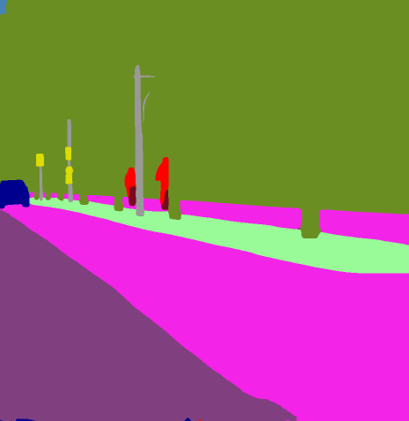} \hspace{-4mm}\\

\end{tabularx}
\caption{\textbf{On Cityscapes \textit{test}:} Showing visual effect of training an HRNet-OCR-HMS model within the HS3 Fuse framework. Notice the improvement in highlighted regions as compared to previous state-of-the-art works on Cityscapes.} \label{fig:city_test_qual}
\end{figure}

\vspace{-5pt}
\subsection{Ablation Studies}
\label{sec:ablation}
\vspace{-5pt}

\noindent \textbf{Finding Optimal Number of Clusters:}
We analyze the segmentation accuracy w.r.t. different performance-complexity trade-offs at the intermediate layers, by varying the parameter of $\theta$. We use an HRNetv2-w18-OCR model on NYUD-v2. Based on the choice of $\theta$, we obtain the numbers of classes $K_1$ and $K_2$ for the first and second intermediate stages. As shown in Table~\ref{tab:theta_ablation}, the network achieves optimal segmentation mIoU of $41.8\%$ when $\theta = 80^{\circ}$. By using our proposed approach of enforcing consistent trade-offs across layers from Section~\ref{sec:right_task_complexity}, we obtain $\theta = 76^{\circ}$. This allows us to achieve a near-optimal mIou of $41.7\%$.

When $\theta=0^\circ$, we recover the vanilla deep supervision which assigns over-complex tasks to intermediate layers. When $\theta=90^\circ$, it is required that these intermediate stages achieve the same mIoU as the final output, which results in over-simplified tasks for them. As shown in Table~\ref{tab:theta_ablation}, both baselines perform considerably worse as compared to our proposed approach.

\vspace{6pt}

\noindent \textbf{Choice of Clustering Methods:}
We study the effect of using other clustering methods within our hierarchical supervision framework: 1) k-means OCR feature clustering and 2) manual assignment. These are mentioned in Section~\ref{sec:alternative_clustering}. We use an HRNetv2-w18 backbone trained using the HS3 scheme and vary our clustering approach. We report our results on the Cityscapes \textit{val} set.

As shown in Table~\ref{tab:clustering_ablation}, HS3 with any of the clustering methods outperforms deep supervision and the case of no auxiliary supervision. Manual assignment under-performs compared to k-means clustering and spectral clustering, as it is based on human intuition and does not properly align with the sub-networks' capabilities. While k-means clustering performs on par with spectral clustering, it requires class-wise embeddings (e.g., the object representations derived in OCR) at each stage. These representations may not be always available in a given network. In contrast, spectral clustering only requires the confusion matrices.



\begin{minipage}[t!]{0.98\textwidth}
\vspace{-8pt}
\hspace{-10pt}
\begin{minipage}[t]{0.51\textwidth}
\makeatletter\def\@captype{table}
\small
  \centering
  \begin{tabular}{c c c| c} 
  \hline
  $\theta$ & $K_1$ & $K_2$ & mIoU\\ 
  \hline \hline
  $0^\circ$ & 40 & 40 & 41.2 \\
  $50^\circ$ & 31 & 34 & 41.2 \\
  $75^\circ$ & 20 & 27 & 41.7 \\
  $80^\circ$ & 17 & 25 & \textbf{41.8} \\
  $85^\circ$ & 10 & 20 & 41.4 \\
  $90^\circ$ & 5 & 16 & 41.5 \\
  \hline
  \end{tabular}
\caption{\small \textbf{On NYUD-v2:} Effect of varying $\theta$ (i.e., trade-off parameter) in HS3 to train HRNetv2-w18-OCR. Our approach derives a near-optimal $\theta=76^{\circ}$ with $41.7\%$ mIoU.}
\label{tab:theta_ablation}
\end{minipage} \hspace{3pt}
\begin{minipage}[t]{0.45\textwidth}
\makeatletter\def\@captype{table}
\small 
 \centering
 \begin{tabular}{c | c | c} 
 \hline
Auxiliary & Clustering &  \multirow{2}{*}{mIoU}\\
Supervision & Method  &\\
 \hline \hline
  None & - & 77.6 \\
  DS & - & 77.7 \\
  HS3 & manual & 77.9 \\
  HS3 & k-means & \textbf{78.1} \\
  HS3 & spectral & \textbf{78.1} \\

  \hline
  \end{tabular}
\caption{\small \textbf{On Cityscapes \textit{val}:} Using various clustering methods with HS3 to train HRNetv2-w18. For k-means, OCR modules are used to extract embeddings.}
\label{tab:clustering_ablation}
\end{minipage}
\end{minipage}

\vspace{-5pt}
\section{Conclusions}
\vspace{-5pt}

In this work, we have presented a training method that supervises the transitional layers of a segmentation network to learn meaningful representations adaptively by varying task complexity. We derived various sets of class clusters to supervise each transitional layer of the network to facilitate this. Furthermore, we devised a fusion framework to leverage additional context offered by our derived hierarchical features. We showed empirically that our proposed training scheme considerably outperforms baselines and also deep supervision with no added inference cost. The proposed fusion architecture offers superior performance on public benchmarks. For future work, we plan to extend our scheme to various tasks, including classification. We're also looking for an acceptable method for single-stage training.

\bibliography{egbib}
\end{document}